# The Stixel World: A Medium-Level Representation of Traffic Scenes

Marius Cordts[a,c,1,*], Timo Rehfeld[b,c,1], Lukas Schneider[a,e,1], David Pfeiffer[a], Markus Enzweiler[a], Stefan Roth[c], Marc Pollefeys[d,e], Uwe Franke[a]

[a]*Department of Environment Perception, Daimler AG, R&D, Sindelfingen, Germany*
[b]*Mercedes-Benz R&D North America, Sunnyvale, CA, USA*
[c]*Department of Computer Science, TU Darmstadt, Germany*
[d]*Microsoft HoloLens, Seattle, WA, USA*
[e]*Institute for Visual Computing, ETH Zurich, Switzerland*

**Abstract**

Recent progress in advanced driver assistance systems and the race towards autonomous vehicles is mainly driven by two factors: (1) increasingly sophisticated algorithms that interpret the environment around the vehicle and react accordingly, and (2) the continuous improvements of sensor technology itself. In terms of cameras, these improvements typically include higher spatial resolution, which as a consequence requires more data to be processed. The trend to add multiple cameras to cover the entire surrounding of the vehicle is not conducive in that matter. At the same time, an increasing number of special purpose algorithms need access to the sensor input data to correctly interpret the various complex situations that can occur, particularly in urban traffic.

By observing those trends, it becomes clear that a key challenge for vision architectures in intelligent vehicles is to share computational resources. We believe this challenge should be faced by introducing a representation of the sensory data that provides compressed and structured access to all relevant visual content of the scene. The Stixel World discussed in this paper is such a representation. It is a medium-level model of the environment that is specifically designed to compress information about obstacles by leveraging the typical layout of outdoor traffic scenes. It has proven useful for a multi-tude of automotive vision applications, including object detection, tracking, segmentation, and mapping.

---

[*]Corresponding author: `marius.cordts@daimler.com`
[1]Authors contributed equally and are listed in alphabetical order

In this paper, we summarize the ideas behind the model and generalize it to take into account multiple dense input streams: the image itself, stereo depth maps, and semantic class probability maps that can be generated, *e.g.*, by deep convolutional neural networks. Our generalization is embedded into a novel mathematical formulation for the Stixel model. We further sketch how the free parameters of the model can be learned using structured SVMs.

## 1. Introduction

In recent years, more and more vision technology has been deployed in vehicles, mainly due to the low cost and versatility of image sensors. As a result, the number of cameras, their spatial resolution, and the list of algorithms that involve image analysis are constantly increasing. This poses significant challenges in terms of processing power, energy consumption, packaging space, and bandwidth, all of which are traditionally limited on embedded hardware commonly found in a vehicle.

In light of these constraints, it becomes clear that an essential part of modern vision architectures in intelligent vehicles is concerned with sharing of resources. One way to share computational resources is outlined by deep neural networks, where dedicated features for specialized tasks share a common set of more generic features in lower levels of the network. This makes deep networks highly attractive for automotive applications. It is, however, challenging to efficiently leverage this information in different computational units in the vehicle while avoiding that each module relying on image information is required to operate on the raw pixel level again. Doing so would significantly increase bandwidth and computational requirements for the underlying system components. One solution to this problem is to find a representation of the image content that abstracts the raw sensory data, while being neither too specific nor too generic, so that it can simultaneously facilitate various tasks such as object detection, tracking, segmentation, localization, and mapping. This representation should allow structured access to depth, semantics, and color information, and should carry high information content while having a low memory footprint to save bandwidth and computational resources.

In a multi-year effort, we have developed the Stixel World [1, 2, 3, 4, 5, 6, 7], a medium-level representation of image and depth data that is specifically designed for the characteristic layout of street scenes. During this time, our model has proven useful for a multitude of practical automotive vision applications. Key observations are that using Stixels as primitive elements either decreases parsing time, increases accuracy, or even both at the same time [8, 9, 10, 11, 12, 13, 14, 15]. Since its introduction, the concept of Stixels has been adopted by several other groups [8, 16, 17, 18, 19] and automotive companies.

Our model is based on the observation that the geometry in man-made environments is dominated by planar surfaces. Moreover, prior knowledge constrains the structure of typical street scenes predominantly in the vertical image direction, *i.e.* objects are on top of a supporting ground plane and the sky is typically in the upper part of the image. Furthermore, there is a characteristic depth ordering from close to far away along this direction. Note that those geometric relations are less pronounced in the horizontal domain. We therefore model the environment as a set of Stixels: thin stick-like elements that constitute a column-wise segmentation of the image, see Fig. 1. Stixels are inferred by solving an energy minimization problem, where structural and semantic prior information is taken into account to regularize the solution. Although we derive and apply our model for the case of image data, the model itself is not limited to this case. It can also be applied to laser range data or other input modalities, as long as the scene layout obeys similar constraints as we observe in street scenes.

In this paper, we extend the existing work on Stixels in several aspects: First, we describe the Stixel model using a novel CRF-based formulation, which encompasses all previously proposed Stixel variants [1, 2, 3, 4, 5, 6, 7] and enables a better comparison to other models in the literature. We then introduce new factors in our graphical model that represent additional input modalities such as color and pixel-level semantic labels. To learn all free parameters of the model, we outline loss-based training using a structured support vector machine (S-SVM). Finally, we provide an in-depth evaluation to analyze the influences of various components and highlight the different aspects of our model. Our experiments are concerned with the inherent properties of the Stixel World itself. Therefore, we do not reiterate the benefits obtained on the application-level through the use of Stixels. For that, we refer to several systems papers that describe and evaluate applications that are built using the Stixel World as underlying representation, *e.g.* [8, 9, 10, 11, 12, 14, 15], including a fully autonomous vehicle with Stixels serving as the underlying visual environment representation [13].

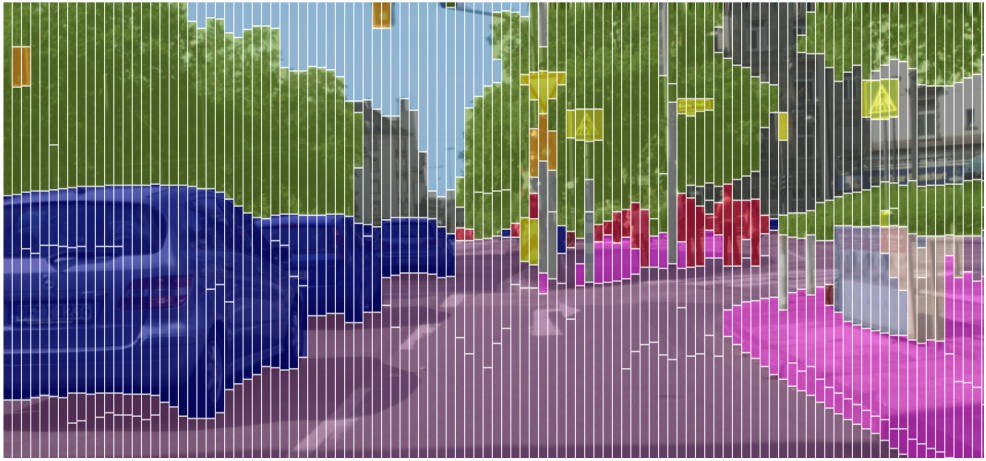

Semantic representation, where Stixel colors encode semantic classes following [20].

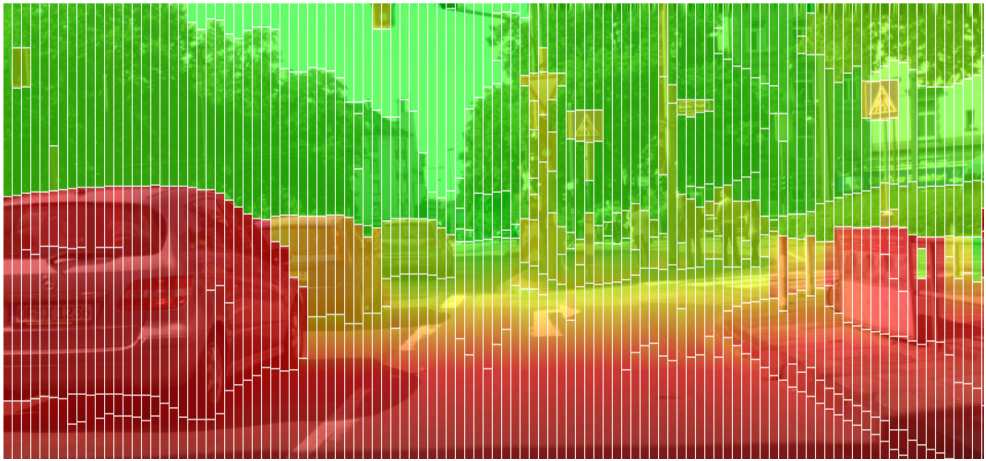

Depth representation, where Stixel colors encode disparities from close (red) to far (green).

Figure 1: Scene representation obtained via Semantic Stixels. The scene is represented via its geometric layout (bottom) and semantic classes (top). Adapted from Schneider *et al.* [7].

## 2. Related Work

Conceptually, our Stixel model can be placed within a triangle of three related lines of research: (1) road scene models, (2) unsupervised bottom-up segmentation, and (3) semantic segmentation. We will proceed to discuss several closely related models within those reference categories.

In the first category of road scene models, occupancy grid maps play a predominant role in representing the surrounding of the vehicle [21, 11, 22, 23]. Such models typically define a grid in bird's eye perspective and reason about the occupancy state of each grid cell to extract the drivable area, obstacles, and unobservable areas from range data. Occupancy grids and the Stixel World are related in that both models represent the 2D image in terms of column-wise stripes. This allows to interpret the camera data in a polar fashion, similar to the data obtained from laser scanners or radar sensors. As a result, the use of Stixels facilitates the integration of image data into an occupancy grid map. Furthermore, the Stixel data model can be regarded as the analog to the forward model typically used in occupancy grid maps, *c.f.* Section 4. However, the Stixel inference scheme in the image plane differs considerably from the methods used in classical grid-based approaches.

The second related category is unsupervised image segmentation, which aims to partition an image into regions of coherent color or texture. Such approaches are both concerned with superpixels [24, 25, 26, 27], as well as larger segments or segmentation hierarchies [28, 29]. These methods typically only have a few model constraints and parameters that govern the segment size. In this sense, they are generic and frequently used as finest level of processing granularity to both increase efficiency and to focus the attention of object detection [30, 31] and semantic segmentation [32, 33]. Stixels are related to these models in that they also provide a compressed representation of the scene that serves as the basis for subsequent processing steps. In contrast to the generality of superpixels, Stixels are more specific in that coarse geometric labels are inferred for each segment by incorporating prior information about the scene layout.

In the domain of semantic segmentation, our third related category, considerable progress has been made by coupling 2D appearance cues with 3D depth information, which is either extracted using multiple cameras or structure-from-motion [34, 35, 36, 37, 38]. In particular, joint segmentation and depth estimation has gained at lot of interest, as depth information provides strong evidence towards certain object classes. At the same time, erroneous or missing depth measurements can be overcome by imposing a specific depth model or smoothness assumption for individual segments [39] or categories [34, 35, 36]. More recently, the resurrection of deep convolutional neural networks – in particular fully convolutional networks (FCNs) [40] – revolutionized segmentation and classification performance. Several methods build on top of FCNs and model statistical dependencies via conditional random fields (CRFs) [41, 42, 43, 44] or incorporate global scene context [45, 46, 47]. In our Stixel model, we also make use of this semantic information and model statistical dependencies on top. However, unlike most previous approaches, Stixels only model one-dimensional constraints to strike a good balance between accuracy and computational complexity in view of real-time automotive vision.

Given this large body of literature, we will now focus on methods that are closely related to Stixels regarding the resulting representation and the underlying inference procedure. In the early work of Hoiem *et al.* [48], the geometric surface layout is extracted from a single image using various features such as location, superpixel shape, color, texture, and vanishing lines. Image segments are classified into the three categories *support*, *vertical*, and *sky*. Segments of the *vertical* class are then subdivided further into coarse surface orientation and property classes such as *solid* and *porous*. In our Stixel model we apply a similar hierarchy of labels, with the same three structural classes (*support*, *vertical*, and *sky*) at the core and a further subdivision into semantic classes within each structural class. The conceptual work from Hoiem *et al.* [48] has later been extended by introducing a method that finds the exact and globally optimal solution to the tiered scene labeling problem [49]. The tiered model assumes that the scene can be segmented vertically into a top, middle, and bottom part, where the middle part can be further subdivided horizontally into a finite set of labels. By exploiting the structure of this layout, the 2D labeling problem is reduced to a 1D problem that can be solved via dynamic programming. While the tiered model in [49] improves quantitatively over [48] in their setting, it is too coarse to resolve the level of detail required to accurately describe complex traffic scenes. Our Stixel model is inspired by the tiered scene labeling idea in terms of imposing a strict geometric model on the scene to reduce the complexity of inference. More recently, Liu *et al.* [17] formulated a layered interpretation of street scenes based on the tiered scene labeling problem and our column-wise Stixel model. They use features from a deep neural network and infer structural and semantic labels jointly, rather than using precomputed depth maps as input. While a joint estimation is desirable in theory, it prohibits pipelining of depth, label, and model computation, which increases the update interval during online inference. Furthermore, their model allows for up to four layers within each column. This is sufficient for a small number of object classes, but does not scale well, if a large number of classes is required to be detected, *c.f.* Fig. 1. Our Stixel formulation is not limited by such constraints as the number of Stixels per column is inferred automatically from the given data. In the following, we introduce our Stixel model and its mathematical formulation in more detail.

## 3. The Stixel Model

The Stixel World $\mathcal{S}$ is a segmentation of an image into superpixels, where each superpixel is a thin stick-like segment with a class label and a 3D planar depth model. The key difference between Stixels and other segmentation approaches is that the segmentation problem of a $w \times h$ image is broken down to $\frac{w}{w_s}$ individual 1D segmentation problems, one for each column of width $w_s$ in the image. The horizontal extent $w_s \geq 1$ is fixed to typically only a few pixels and chosen in advance to reduce the computational complexity during inference. The vertical extent of each Stixel is inferred explicitly in our model. To further control the runtime of our method, we apply an optional downscaling in the vertical direction by a factor of $h_s \geq 1$.

The idea behind the proposed approach is that the dominant structure in road scenes occurs in the vertical domain and can thus be modeled without taking horizontal neighborhoods into account. This simplification allows for efficient inference, as all columns can be segmented in parallel. We regard our model as a *medium-level* representation for three reasons: (1) each Stixel provides an abstract representation of depth, physical extent, and semantics that is more expressive than individual pixels; (2) the Stixel segmentation is based upon a street scene model, compared to bottom-up super-pixel methods; (3) Stixels are not a high-level representation, as individual object instances are covered by multiple Stixels in their horizontal extent. All in all, Stixels deliver a compressed scene representation that subsequent higher-level processing stages can build on.

Our label set is comprised of three *structural classes*, "support" ($\mathcal{S}$), "vertical" ($\mathcal{V}$), and "sky" ($\mathcal{Y}$). All three structural classes can be distinguished exclusively by their geometry and reflect our underlying 3D scene model: support Stixels are parallel to the ground plane at a constant height and ver-tical/sky Stixels are perpendicular to the ground plane at a constant/infinite distance. We further refine the structural classes to *semantic classes* that are mapped onto the sets $\mathcal{S}$, $\mathcal{V}$, or $\mathcal{Y}$. Semantic classes such as road or sidewalk, for example, are in the support set $\mathcal{S}$, whereas building, tree or vehicle are vertical, *i.e.* in $\mathcal{V}$. The actual set of semantic classes is highly dependent on the application and is further discussed in the experiments in Section 7.

The input data for our inference is comprised of a dense depth map, a color image, and pixel-level label scores for each semantic class. Note that all three input channels are optional and are not restricted to specific stereo or pixel classification algorithms. In the following sections, we focus on a single image column of width $w_s$ and provide a detailed mathematical description of the Stixel model. Our graphical model defines the energy function and gives an intuition on factorization properties and thus statistical independence assumptions. Further, we describe an efficient energy minimization procedure via dynamic programming (DP) yielding the segmentation of one image column. Subsequently, we sketch learning of the model parameters from ground truth data via structural support vector machines (S-SVMs).

## 4. Graphical Model

In the following, we define the posterior distribution $P(\boldsymbol{S}_: \mid \boldsymbol{M}_:)$ of a Stixel column $\boldsymbol{S}_:$ given the measurements $\boldsymbol{M}_:$ within the column. These measurements consist of a dense disparity map $\boldsymbol{D}_:$, the color image $\boldsymbol{I}_:$, and pixel-level semantic label scores $\boldsymbol{L}_:$. We describe the posterior by means of a graphical model to allow for an intuitive and clear formalism, a structured description of statistical independence, and a visualization via a factor graph, *c.f.* Fig. 2. The graph provides a factorization grouped into a likelihood $\tilde{P}(\boldsymbol{M}_: \mid \boldsymbol{S}_:)$ and a prior $\tilde{P}(\boldsymbol{S}_:)$, giving

$$P(\boldsymbol{S}_: \mid \boldsymbol{M}_:) = \frac{1}{Z} \tilde{P}(\boldsymbol{M}_: \mid \boldsymbol{S}_:) \tilde{P}(\boldsymbol{S}_:) \ , \qquad (1)$$

where $Z$ is the normalizing partition function. Note that both, likelihood and prior are unnormalized probability mass functions. Switching to the log-domain, we obtain

$$P(\boldsymbol{S}_: = \boldsymbol{s}_: \mid \boldsymbol{M}_: = \boldsymbol{m}_:) = e^{-E(\boldsymbol{s}_:, \boldsymbol{m}_:)} \ , \qquad (2)$$

where $E(\cdot)$ is the energy function, defined as

$$E(\boldsymbol{s}_:, \boldsymbol{m}_:) = \Phi(\boldsymbol{s}_:, \boldsymbol{m}_:) + \Psi(\boldsymbol{s}_:) - \log(Z) \ . \qquad (3)$$

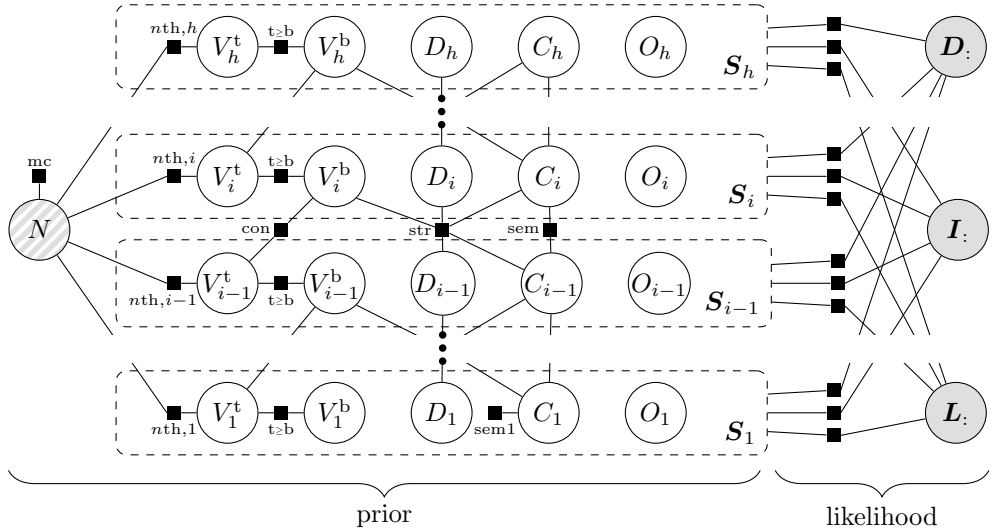

Figure 2: The Stixel world as a factor graph that depicts the factorization of the posterior distribution, *c.f.* Eq. (1). Each Stixel $S_i$ (dashed boxes) stands for the five random variables $V_i^t, V_i^b, C_i, D_i, O_i$ (circles). The hatched node on the left denotes the random variable $N$ describing the number of Stixels $n$ that constitute the final segmentation. Black squares denote factors of the posterior and are labeled according to the descriptions given in the text. The prior distribution factorizes according to the left part, whereas the right part describes the measurement likelihood. The circles $\boldsymbol{D}_:$, $\boldsymbol{I}_:$, and $\boldsymbol{L}_:$ denote the measurements, *i.e.* a column of the disparity map, the color image, and the pixel-level semantic label scores, respectively. If all measurements are observed (indicated by gray shading) and if the number of Stixels $N$ is thought to be fixed (indicated by gray hatched shading), the graph is chain-structured on the Stixel level. This property is exploited for inference via dynamic programming (see Section 5).

The function $\Phi(\cdot)$ represents the likelihood and $\Psi(\cdot)$ the prior.

In order to mathematically define the posterior energy function, we need a fixed number of random variables describing the segmentation $\boldsymbol{S}_:$. This is accomplished by splitting the column $\boldsymbol{S}_:$ into as many individual Stixels $\boldsymbol{S}_i$ as maximally possible, *i.e.* $i \in \{1\ldots h\}$, and into an additional random variable $N \in \{1\ldots h\}$ denoting the number of Stixels that constitute the final segmentation. For a certain value $n$, we set all factors connected to a Stixel $\boldsymbol{S}_i$ with $i > n$ to zero energy. Thus, these factors do not influence the segmentation obtained. For ease of notation, we continue by assuming $i \leq n$ and drop the dependency of the factors on $N$. We revisit these simplifications and discuss their impact during inference later in Section 5.

Besides the random variable $N$, the column segmentation $\boldsymbol{S}_:$ consists of $h$ Stixels $\boldsymbol{S}_i$, where $\boldsymbol{S}_1$ is the lowest Stixel and $\boldsymbol{S}_h$ the highest. A Stixel $\boldsymbol{S}_i$ in turn splits into the five random variables $V_i^{\text{b}}$, $V_i^{\text{t}}$, $C_i$, $O_i$, and $D_i$. The first two denote the vertical extent from bottom to top row. The variable $C_i$ represents the Stixel's semantic class label, $O_i$ is its color attribute, and $D_i$ parameterizes the disparity model. Note that a vertical segment at constant distance maps to a constant disparity, while a support segment at constant height maps to a constant disparity offset relative to the ground plane.

### 4.1. Prior

The prior $\Psi(\boldsymbol{s}_:)$ from Eq. (3) captures prior knowledge on the scene structure independent from any measurements and factorizes as

$$\Psi(\boldsymbol{s}_:) = \Psi_{\text{mc}}(n) + \sum_{i=1}^{h} \sum_{\text{id}} \Psi_{\text{id}}(\boldsymbol{s}_i, \boldsymbol{s}_{i-1}, n) \;, \tag{4}$$

where *id* stands for the name of the factors corresponding to their labels in Fig. 2. Note that not all factors actually depend on all variables $\boldsymbol{s}_i$, $\boldsymbol{s}_{i-1}$, or $n$. In the following, we define and explain the individual factors by grouping them regarding their functionality, *i.e.* model complexity, segmentation consistency, structural priors, and semantic priors.

**Model complexity.** The model complexity prior $\Psi_{\text{mc}}$ is the main regularization term and controls the compromise between compactness and robustness versus fine granularity and accuracy. The factor is defined as

$$\Psi_{\text{mc}}(n) = \beta_{\text{mc}}\, n \;. \tag{5}$$

The higher the parameter $\beta_{\mathrm{mc}}$ is chosen, the fewer Stixels are obtained, hence the segmentation becomes more compact.

**Segmentation consistency.** In order to obtain a consistent segmentation, we define hard constraints to satisfy that segments are non-overlapping, connected, and extend over the whole image. This implies that the first Stixel must begin in image row 1 (bottom row) and the last Stixel must end in row $h$ (top row), *i.e.*

$$\Psi_{\mathrm{1st}}(v_1^{\mathrm{b}}) = \begin{cases} 0 & \text{if } v_1^{\mathrm{b}} = 1 \\ \infty & \text{otherwise} \end{cases}, \tag{6}$$

$$\Psi_{\mathrm{nth},i}(n, v_i^{\mathrm{t}}) = \begin{cases} \infty & \text{if } n = i \text{ and } v_i^{\mathrm{t}} \neq h \\ 0 & \text{otherwise} \end{cases}. \tag{7}$$

Further, a Stixel's top row must be above the bottom row and consecutive Stixels must be connected, *i.e.*

$$\Psi_{\mathrm{t\geq b}}(v_i^{\mathrm{b}}, v_i^{\mathrm{t}}) = \begin{cases} 0 & \text{if } v_i^{\mathrm{b}} \leq v_i^{\mathrm{t}} \\ \infty & \text{otherwise} \end{cases}, \tag{8}$$

$$\Psi_{\mathrm{con}}(v_i^{\mathrm{b}}, v_{i-1}^{\mathrm{t}}) = \begin{cases} 0 & \text{if } v_i^{\mathrm{b}} = v_{i-1}^{\mathrm{t}} + 1 \\ \infty & \text{otherwise} \end{cases}. \tag{9}$$

In Section 5, we will show that these deterministic constraints can be exploited to significantly reduce the computational effort.

**Structural priors.** Road scenes have a typical 3D layout in terms of the structural classes support, vertical, and sky. This layout is modeled by $\Psi_{\mathrm{str}}$, which is in turn comprised of two factors, both responsible for an individual effect, *i.e.*

$$\Psi_{\mathrm{str}}(c_i, c_{i-1}, d_i, d_{i-1}, v_i^{\mathrm{b}}) = \Psi_{\mathrm{grav}} + \Psi_{\mathrm{do}}. \tag{10}$$

The gravity component $\Psi_{\mathrm{grav}}$ is only non-zero for $c_i \in \mathcal{V}$ and $c_{i-1} \in \mathcal{S}$ and models that in 3D a vertical segment usually stands on the preceding support surface. Consequently, the vertical segment's disparity $d_i$ and the disparity of the support surface must coincide in row $v_i^{\mathrm{b}}$. The latter disparity is denoted by the function $d_s(v_i^{\mathrm{b}}, d_{i-1})$ and their difference as $\Delta_d = d_i - d_s(v_i^{\mathrm{b}}, d_{i-1})$.

Then, $\Psi_{\text{grav}}$ is defined as

$$\Psi_{\text{grav}} = \begin{cases} \alpha_{\text{grav}}^- + \beta_{\text{grav}}^- \Delta_d & \text{if } \Delta_d < 0 \\ \alpha_{\text{grav}}^+ + \beta_{\text{grav}}^+ \Delta_d & \text{if } \Delta_d > 0 \\ 0 & \text{otherwise} \end{cases} . \quad (11)$$

The second term $\Psi_{\text{do}}$ rates the depth ordering of vertical segments. Usually, an object that is located on top of another object in the image is behind the other one in the 3D scene, *i.e.* the top disparity is smaller than the bottom one. Therefore, we define

$$\Psi_{\text{do}} = \begin{cases} \alpha_{\text{do}} + \beta_{\text{do}}(d_i - d_{i-1}) & \text{if } c_i \in \mathcal{V}, c_{i-1} \in \mathcal{V}, d_i > d_{i-1} \\ 0 & \text{otherwise} \end{cases} . \quad (12)$$

**Semantic priors.** The last group of prior factors is responsible for the a-priori knowledge regarding the semantic structure of road scenes. We define the factor as

$$\Psi_{\text{sem}}(c_i, c_{i-1}) = \gamma_{c_i} + \gamma_{c_i, c_{i-1}} . \quad (13)$$

The first term $\gamma_{c_i}$ describes the a-priori class probability for a certain class $c_i$, *i.e.* the higher the values are, the less likely are Stixels with that class label. The latter term $\gamma_{c_i, c_{i-1}}$ is defined via a two-dimensional transition matrix $\gamma_{c_i, c_{i-1}}$ for all combinations of classes. Individual entries in this matrix model expectations on relative class locations, *e.g.* a support Stixel such as road above a vertical Stixel such as car might be rated less likely than vice versa. Note that we capture only first order relations to allow for efficient inference. Finally, we define $\Psi_{\text{sem1}}(c_1)$ analogously for the first Stixel's class, *e.g.* a first Stixel with a support class such as road might be more likely than with a vertical class such as infrastructure or sky.

*4.2. Data likelihood*

The data likelihood from Eq. (3) integrates the information from our input modalities (disparity map $\boldsymbol{D}_:$, color image $\boldsymbol{I}_:$, and semantic label scores $\boldsymbol{L}_:$) and factorizes as

$$\Phi(\boldsymbol{s}_:, \boldsymbol{m}_:) = \sum_{i=1}^{h} \sum_{v=v_i^{\text{b}}}^{v_i^{\text{t}}} \Phi_D(\boldsymbol{s}_i, d_v, v) + \Phi_I(\boldsymbol{s}_i, i_v) + \Phi_L(\boldsymbol{s}_i, \boldsymbol{l}_v) . \quad (14)$$

Note that we sum over the maximum number of Stixels $h$, but as described above, all factors for $i > n$ are set to zero. Further, for a given Stixel segmentation $\boldsymbol{s}_{:}$, we model the data likelihoods to be independent across pixels and therefore their contribution decomposes over the rows $v$. In the following, we describe the contributions of the individual modalities to a Stixel $\boldsymbol{s}_i$.

**Depth.** Leveraging depth information allows to separate support from vertical segments as well as multiple stacked objects at different distances. The depth likelihood terms are designed according to our world model consisting of supporting and vertical planar surfaces. Since the width $w$ of a Stixel is rather small, we can neglect the influence of slanted surfaces. Instead, these are represented, with some discretization, via neighboring Stixels at varying depths. In doing so, the 3D orientation of a Stixel is sufficiently described by its structural class, *i.e.* support or vertical. Accordingly, the 3D position of a Stixel is parametrized by a single variable $D_v$ paired with its 2D position in the image. This variable is the Stixel's constant disparity for a vertical segment and a constant disparity offset relative to the ground plane for a support segment, *c.f.* Fig. 3.

We use depth measurements in the form of dense disparity maps, where each pixel has an associated disparity value or is flagged as invalid, *i.e.* $d_v \in \{0 \ldots d_{\max}, d_{\text{inv}}\}$. The subscript $v$ denotes the row index within the considered column. The depth likelihood term $\Phi_D(\boldsymbol{s}_i, d_v, v)$ is derived from a probabilistic, generative measurement model $P_v(D_v = d_v \mid \boldsymbol{S}_i = \boldsymbol{s}_i)$ according to

$$\Phi_D(\boldsymbol{s}_i, d_v, v) = -\delta_D(c_i) \log(P_v(D_v = d_v \mid \boldsymbol{S}_i = \boldsymbol{s}_i)) \ . \tag{15}$$

The term incorporates class-specific weights $\delta_D(c_i)$ that allow to learn the relevance of the depth information for a certain class. Let $p_{\text{val}}$ be the prior probability of a valid disparity measurement. Then, we obtain

$$P_v(D_v \mid \boldsymbol{S}_i) = \begin{cases} p_{\text{val}} \ P_{v,\text{val}}(D_v \mid \boldsymbol{S}_i) & \text{if } d_v \neq d_{\text{inv}} \\ (1 - p_{\text{val}}) & \text{otherwise} \end{cases}, \tag{16}$$

where $P_{v,\text{val}}(D_v \mid \boldsymbol{S}_i)$ denotes the measurement model of valid measurements only and is defined as

$$P_{v,\text{val}}(D_v \mid \boldsymbol{S}_i) = \frac{p_{\text{out}}}{Z_U} + \frac{1 - p_{\text{out}}}{Z_G(\boldsymbol{s}_i)} e^{-\frac{1}{2}\left(\frac{d_v - \mu(\boldsymbol{s}_i, v)}{\sigma(\boldsymbol{s}_i)}\right)^2} \ . \tag{17}$$

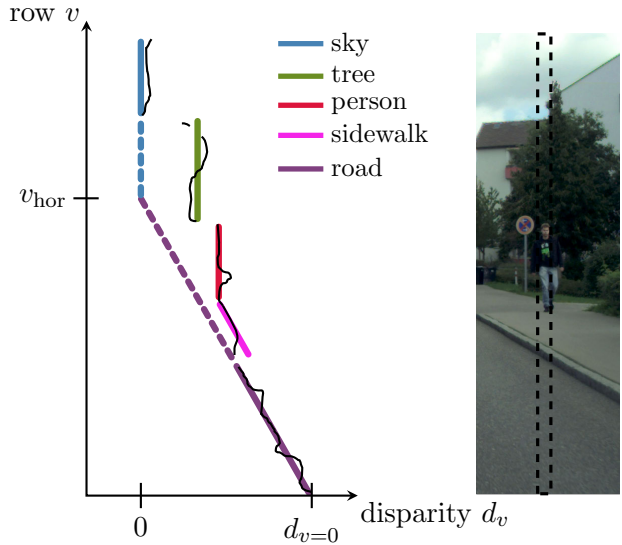

Figure 3: Disparity measurements (black lines) and resulting Stixel segmentation (colored lines) for a typical image column (right). The purple dashed line symbolizes the linear ideal disparity measurements along a planar ground surface that support segments such as road or sidewalk are parallel to. Obstacles form vertical Stixels, *e.g.* person or tree. Sky is modeled with a disparity value of zero. Adapted from Pfeiffer *et al.* [3].

This distribution is a mixture of a uniform and a Gaussian distribution and defines the sensor model. While the Gaussian captures typical disparity measurement noise, the uniform distribution increases the robustness against outliers and is weighted with the coefficient $p_{\text{out}}$. The Gaussian is centered at the disparity value $\mu(\boldsymbol{s}_i, v)$ of the Stixel $\boldsymbol{s}_i$, which is constant for vertical Stixels, *i.e.* $\mu(\boldsymbol{s}_i, v) = d_i$, and depends on row $v$ for support Stixels, *c.f.* Fig. 3. The standard deviation $\sigma$ captures the noise properties of the stereo algorithm and is chosen depending on the class $c_i$, *e.g.* for class sky the noise is expected to be higher than for the other classes due to missing texture as needed by stereo matching algorithms. The parameters $p_{\text{val}}$, $p_{\text{out}}$, and $\sigma$ are either chosen a-priori, or can be obtained by estimating confidences in the stereo matching algorithm as shown in [4]. Later in Section 6, we further show how to choose these parameters based on empirical results. The terms $Z_G(\boldsymbol{s}_i)$ and $Z_U$ normalize the two distributions.

**Color image.** Common superpixel algorithms such as SLIC [24] work by grouping adjacent pixels of similar color. We follow this idea by favoring Stixels with a small squared deviation in LAB color space from their color attribute $O_i$. Thus, the color likelihood $\Phi_I(\boldsymbol{s}_i, i_v)$ in Eq. (14) is defined as

$$\Phi_I(\boldsymbol{s}_i, i_v) = \delta_I(c_i) \|i_v - o_i\|_2^2 \ . \tag{18}$$

Since some classes might satisfy a constant color assumption better than others, *e.g.* sky *vs.* object, each class is associated with an individual weight $\delta_I$. Note that it is straightforward to extend this likelihood term to more sophisticated color or texture models. In this work, we opt for semantic label scores to incorporate appearance information.

**Semantic label scores.** The driving force in terms of semantic scene information is provided by a pixel-level labeling system that delivers normalized semantic scores $l_v(c_i)$ with $\sum_{c_i} l_v(c_i) = 1$ for all considered classes $c_i$ at all pixels $v$. This input channel not only separates the structural classes into their subordinate semantic classes, but also guides the segmentation by leveraging appearance information. The semantic scores yield the likelihood term

$$\Phi_L(\boldsymbol{s}_i, \boldsymbol{l}_v) = -\delta_L(c_i) \log(l_v(c_i)) \tag{19}$$

in Eq. (14). Again, we use class-specific weights $\delta_L(c_i)$.

## 5. Inference

We perform inference by finding the maximum-a-posteriori (MAP) solution by maximizing Eq. (1), or equivalently by minimizing the energy function in Eq. (3). One motivation for this is that as opposed to a maximum marginal estimate, the obtained segmentation is consistent in terms of the constraints described in Section 4.1. We describe the inference algorithm in three stages: first using a naive solution, then by exploiting algorithmic simplifications, and eventually by using slight approximations to further reduce the computational effort.

### 5.1. Dynamic programming

If we treat the given measurements as implicit parameters and combine Eqs. (3), (4) and (14), the optimization problem has the structure

$$\boldsymbol{s}_:^\star = \underset{n, \boldsymbol{s}_1 \ldots \boldsymbol{s}_h}{\operatorname{argmin}} \ \Psi_{\mathrm{mc}}(n) + \Phi_1(\boldsymbol{s}_1, n) + \Psi_1(\boldsymbol{s}_1, n) + \sum_{i=2}^{h} \Phi_i(\boldsymbol{s}_i, n) + \Psi_i(\boldsymbol{s}_i, \boldsymbol{s}_{i-1}, n) \ , \tag{20}$$

where all prior and likelihood factors of a single Stixel $i$ or a pair of neighboring Stixels are grouped together. A straightforward way to solve Eq. (20) is to iterate over all possible numbers of Stixels $n$ and solve

$$c_n^\star = \Psi_{\mathrm{mc}}(n) + \min_{\boldsymbol{s}_1\ldots\boldsymbol{s}_h}\Phi_{1,n}(\boldsymbol{s}_1) + \Psi_{1,n}(\boldsymbol{s}_1) + \sum_{i=2}^{h}\Phi_{i,n}(\boldsymbol{s}_i) + \Psi_{i,n}(\boldsymbol{s}_i, \boldsymbol{s}_{i-1}) \quad (21)$$

for each fixed $n$. The minimum value of all $c_n^\star$ determines $n^\star$ and the minimizing segmentation $\boldsymbol{s}_:^\star$ is the optimal segmentation of the current column. Further, for a fixed $n$, we exploit that $\Phi_{i,n} = \Psi_{i,n} = 0$ for $i > n$ and that the factor $\Psi_{n\mathrm{th},i}$ reduces to the constraint $v_n^\mathrm{t} = h$. Besides that, neither $\Phi_{i,n}$ nor $\Psi_{i,n}$ depend on $n$ or $i$ (except for $i = 1$) and it holds that

$$c_n^\star = \Psi_{\mathrm{mc}}(n) + \min_{\substack{\boldsymbol{s}_1\ldots\boldsymbol{s}_n \\ v_n^\mathrm{t}=h}} \Phi(\boldsymbol{s}_1) + \Psi_1(\boldsymbol{s}_1) + \sum_{i=2}^{n}\Phi(\boldsymbol{s}_i) + \Psi(\boldsymbol{s}_i, \boldsymbol{s}_{i-1}) \ . \quad (22)$$

Due to the first-order Markov property on Stixel super-nodes, *c.f.* Fig. 2, Eq. (22) can be solved via the Viterbi algorithm, *i.e.* dynamic programming, by reformulation as

$$\begin{aligned}
c_n^\star = \Psi_{\mathrm{mc}}(n) + &\min_{\boldsymbol{s}_n, v_n^\mathrm{t}=h} \Phi(\boldsymbol{s}_n) + \\
&\min_{\boldsymbol{s}_{n-1}} \Phi(\boldsymbol{s}_{n-1}) + \Psi(\boldsymbol{s}_n, \boldsymbol{s}_{n-1}) + \\
&\vdots \\
&\min_{\boldsymbol{s}_1} \Phi(\boldsymbol{s}_1) + \Psi(\boldsymbol{s}_2, \boldsymbol{s}_1) + \Psi_1(\boldsymbol{s}_1) \ .
\end{aligned} \quad (23)$$

The number of possible states of a Stixel $\boldsymbol{S}_i$ is

$$|\boldsymbol{S}_i| = |V^\mathrm{t}||V^\mathrm{b}||C||O||D| = h^2 |C||O||D| \ , \quad (24)$$

and hence we obtain a run time of $\mathcal{O}(N|\boldsymbol{S}_i|^2) = \mathcal{O}(Nh^4|C|^2|O|^2|D|^2)$ for each value of $n$. Since inference is run for $n \in \{1\ldots h\}$, the overall run time is $\mathcal{O}(h^6|C|^2|O|^2|D|^2)$. This assumes that all prior and likelihood factors can be computed in constant time. In case of the priors, this property becomes evident from the definitions in Section 4.1. For the data likelihoods, constant run time is achieved by leveraging integral tables within each image column. For the disparity data term, we apply the approximations described in [3] to compute the integral table of the disparity measurements. Overall, the asymptotic run time of such pre-computations is small compared to the inference via the Viterbi algorithm and can thus be neglected.

## 5.2. Algorithmic simplification

The run time can be significantly reduced by exploiting the structure of the optimization problem in Eq. (23). All intermediate problems neither depend on the number of Stixels $n$ nor on the actual Stixel index $i$, except for $i = 1$. Further, the model complexity factor $\Psi_{\mathrm{mc}}(n)$ is linear in the number of Stixels, *c.f.* Eq. (5), and can thus be transformed into a constant unary term for each Stixel. Therefore, inference can be performed jointly for all values of $n$ and the overall run time reduces to $\mathcal{O}(|\boldsymbol{S}_i|^2) = \mathcal{O}(h^4 |C|^2 |O|^2 |D|^2)$.

Further improvement is obtained by exploiting the deterministic constraints on the segmentation consistency, *c.f.* Section 4.1. The random variable $V_i^{\mathrm{b}}$ can be substituted with $V_{i-1}^{\mathrm{t}} + 1$ for $i > 1$ and with 1 for $i = 1$, *c.f.* the factors $\Psi_{\mathrm{con}}$ and $\Psi_{\mathrm{1st}}$ in Eqs. (6) and (9). As shown in Fig. 2, there is no connection between $V_i^{\mathrm{b}}$ and any random variable from Stixel $\boldsymbol{S}_{i+1}$. Thus, the substitution does not add a second-order dependency and an inference via the Viterbi algorithm is still possible. However, the number of states is reduced to

$$|\boldsymbol{S}_i| = |V^{\mathrm{t}}| |C| |O| |D| = h |C| |O| |D| \ , \tag{25}$$

and the overall run time becomes $\mathcal{O}(h^2 |C|^2 |O|^2 |D|^2)$.

An inspection of all factors defined in Section 4 unveils that no pairwise factor depends on the Stixel's color attribute $O_i$. Instead, only the image data likelihood in Eq. (18) depends on $O_i$ and its minimization is solved analytically, *i.e.* $o_i^\star$ is the mean image within the Stixel $\boldsymbol{s}_i$. Both, $o_i^\star$ and the value of Eq. (18), can be computed in constant time using integral tables. The run time is therefore reduced to $\mathcal{O}(h^2 |C|^2 |D|^2)$.

Let us now discuss the role of $C_i$, *i.e.* the Stixel's semantic class and indirectly its structural class. The structural prior in Eq. (10) depends only on the structural class, but not on the semantic class, while the semantic prior in Eq. (13) can in principle model transitions between semantic classes. However, in practice it is sufficient to restrict this prior to structural classes only and add exceptions for a few select classes depending on the actual application, *e.g.* [5]. The same holds for the weights of the data likelihoods, *i.e.* $\delta_D(c_i)$ , $\delta_I(c_i)$ , $\delta_L(c_i)$. Since there is a fixed number of three structural classes referring to our world model, the evaluation of all pairwise terms is constant with respect to the number of classes. Only the data likelihood in Eq. (19) depends on $|C|$, yielding a run time of $\mathcal{O}(h^2 |C| |D|^2)$.

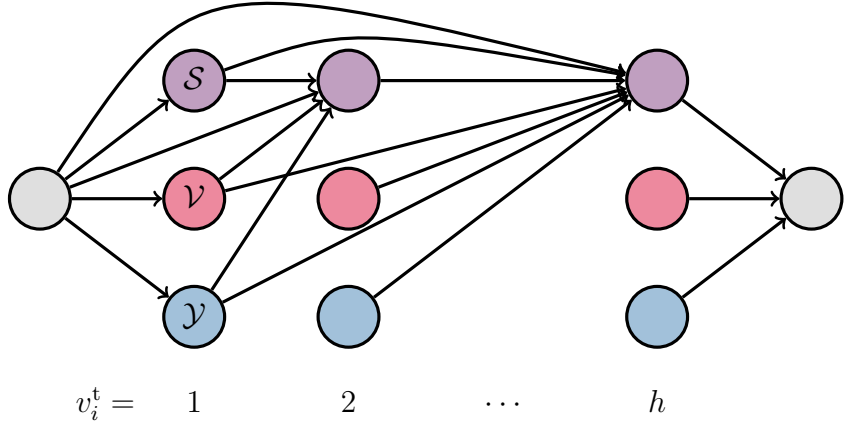

Figure 4: Stixel inference algorithm as a shortest path problem, where the Stixel segmentation is obtained by the colored nodes along the shortest path from the source (left gray node) to the sink (right gray node). The color of the circles denotes the Stixel's structural class, *i.e.* support (purple), vertical (red), and sky (blue). The horizontal position of the nodes is the Stixel's top row $v_i^t$. The graph is a directed acyclic graph (DAG), where only the incoming edges of support nodes are shown.

*5.3. Approximations*

To further reduce the inference effort, we do not infer the variables $D_i$ that describe the Stixels' 3D positions. Instead, the value is computed from the given disparity map depending on the Stixel's extent and structural class label by averaging the disparities within a vertical Stixel respectively the disparity offsets within a support Stixel. In doing so, the isolated disparity data likelihood terms are minimized individually, but the global minimum including the priors is not found exactly. Note that the Stixel's 3D position still depends on its extent, which in turn is inferred by solving the global minimization problem. The average disparities can be computed in constant time using integral tables and the overall run time reduces to $\mathcal{O}(h^2 \left|C\right|)$, which is a significant improvement compared to the initial run time of $\mathcal{O}\bigl(h^6 \left|C\right|^2 \left|O\right|^2 \left|D\right|^2\bigr)$.

A different interpretation of the inference problem is to find the shortest path in the directed acyclic graph in Fig. 4. The edge weights in that graph are defined according to Eq. (22). For an edge between two Stixels $\boldsymbol{s}_l$ (left) and $\boldsymbol{s}_r$ (right), the weight is $\Phi(\boldsymbol{s}_l) + \Psi(\boldsymbol{s}_r, \boldsymbol{s}_l)$. Between the source and a Stixel $s_r$, the weight is $\Psi_1(s_r)$ and between a Stixel $s_l$ and the sink we obtain $\Phi(s_l)$. Note that in all three cases, the model complexity factor is decomposed over these weights, yielding an additional term $\beta_{\mathrm{mc}}$. Solving the shortest path problem is bound by the number of edges $\mathcal{O}(h^2)$ and finding the optimal semantic class for each edge in $\mathcal{O}(|C|)$, yielding $\mathcal{O}(h^2 |C|)$.

## 6. Parameter Learning

The Stixel segmentation is obtained by minimizing the energy function as introduced in Section 4, which is in principle controlled by two groups of parameters. The first group holds the parameters of the generative disparity model, which accounts for the planar Stixel assumption as well as the measurement uncertainty of depth estimates, *i.e.* $p_{\mathrm{val}}$, $p_{\mathrm{out}}$, and $\sigma$ in Eq. (16) and Eq. (17). To derive these parameters from data we use the Cityscapes dataset [20], where we compute depth maps with the semi-global matching (SGM) stereo algorithm [50, 51]. For the prior of valid disparity measurements $p_{\mathrm{val}}$, we count the relation of valid and invalid disparities, yielding $p_{\mathrm{val}} = 0.92$. To find a good estimate of $p_{\mathrm{out}}$ and $\sigma$, we generate ground truth Stixels by cutting instance-level annotation images from the Cityscapes dataset into individual columns and assigning the corresponding ground truth class label to every instance segment. In this experiment, we limit ourselves to vertical Stixels and assign the median of all SGM depth measurements as approximate ground truth depth, assuming that this approximation is sufficient to estimate the parameters we are interested in.

In Fig. 5, we plot the empirical distribution of measurements $d_v$ around the Stixel disparity hypothesis function $\mu(s_i, v)$ for some selected semantic classes in the vertical category. We plot in logarithmic scale to better highlight the nature of the data, with a dense accumulation of points around zero and a heavy tail-like distribution of measurements that are far away from the expected value. It is important to note that these curves capture both the disparity estimation error as well as the error of the Stixel model assumption of obstacles at constant depth. Most interestingly, a single Gaussian fitted to the data in Fig. 5 has a standard deviation $\sigma$ close to 0.5, which is roughly the estimation uncertainty of the SGM algorithm. This strongly indicates that the constant depth model assumption of Stixels is indeed a sensible choice for outdoor traffic scenarios, albeit not always satisfied perfectly, as can be seen from the asymmetric curves, in particular for cars and vegetation. Based on the results in Fig. 5, we choose the two parameters in $P_{v,\text{val}}(D_v \mid \boldsymbol{S}_i)$ as $p_{\text{out}} = 0.01$ and $\sigma = 0.5$, which is a good general approximation of the measured distributions given our model in Eq. (17). We choose $p_{\text{out}}$ conservatively to allow for more severe outliers that might not have been reflected accurately in the dataset given that it was recorded in fair weather conditions.

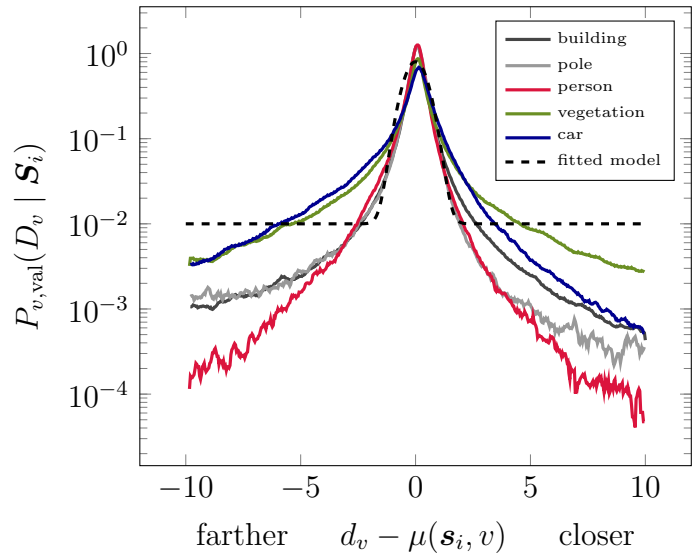

Figure 5: Empirical distribution of measurements around the Stixel disparity hypothesis function, estimated based on ground truth Stixels generated on the Cityscapes dataset [20]. The dashed curve represents the actual Stixel model in Eq. (17), fitted to best represent the measured data distribution, with $p_{\text{out}} = 0.01$ and $\sigma = 0.5$.

The energy function is linear in all remaining parameters forming the second group of parameters, i.e. model complexity ($\beta_{\text{mc}}$), structural priors ($\alpha_{\text{grav}}^-$, $\beta_{\text{grav}}^-$, $\alpha_{\text{grav}}^+$, $\beta_{\text{grav}}^+$, $\alpha_{\text{do}}$, $\beta_{\text{do}}$), semantic priors ($\gamma_{c_i}$, $\gamma_{c_i,c_{i-1}}$) and the weights between the data likelihoods ($\delta_D(c_i)$, $\delta_I(c_i)$, $\delta_L(c_i)$). As tuning these parameters manually can be a tedious task, we investigate automatic parameter learning by means of a structured support vector machine (S-SVM) with margin-rescaled Hinge loss [52]. To do so, we generate ground truth Stixels as described above and define a loss that compares a ground truth Stixel segmentation with an inferred one. We define the loss in a way that it is additive over individual inferred Stixels and punishes over- and under-segmentation as well as a wrong structural class. In doing so, the loss is simply a fourth data term, hence the inference during training is tractable using the same method and approximations as in Section 5. While the sketched structured learning approach works in principle and converges to a meaningful parameter set, we found that the obtained results are comparable, but never surpass manual tuning that is guided by empirical observations of street scenes. We experimented with different loss variants, but in all cases manually tuned parameters were still slightly better, even in terms of the loss used for training. We attribute this effect to the convex relaxation of the Hinge loss, which in our case may not be tight enough. Hence, parameter learning for the Stixel model yields a valid starting point, but further manual guidance by physical evidence helps to increase robustness and generalization abilities.

## 7. Experiments

In this paper, we focus on evaluating the Stixel model by performing validation experiments that provide insights into the influence of different input modalities, model assumptions, core parameters, and computational approximations w.r.t. the accuracy of disparity and semantic label estimation. Throughout all experiments, we use SGM [50] and FCN [40] to obtain dense disparity and semantic label estimates, respectively. For training the neural network, we follow [7]. The following description of the evaluation datasets (Section 7.1) and metrics (Section 7.2) are adapted from Schneider *et al.* [7].

### 7.1. Datasets

From the small number of available realistic outdoor datasets in the area of autonomous driving, the subset of KITTI [53] annotated by Ladicky *et al.* [54] is, to the best of our knowledge, the only dataset containing dense semantic labels and depth ground truth. Therefore, this is the only dataset that allows to report performance metrics of both aspects of our Semantic Stixel representation on the same dataset. It consists of 60 images with a resolution of 0.5 MP, which we entirely use for evaluation; no images are used for training. We follow the suggestion of Ladicky *et al.* to ignore the three rarest object classes, leaving a set of 8 classes.

As a second dataset, we report disparity performance on the training data of the stereo challenge in KITTI'15 [55]. This dataset comprises a set of 200 images with sparse disparity ground truth obtained from a Velodyne HDL-64 laser scanner. However, there is no suitable semantic ground truth available for this dataset.

Third, we evaluate on Cityscapes [20], a recently proposed highly complex and challenging dataset with dense annotations of 19 classes on 2975 images for training and 500 images for validation that we used for testing. While there are stereo views available, ground-truth disparities do not exist.

### 7.2. Metrics

In our experiments, we use four different metrics that are designed to assess the viability of our Semantic Stixel model and several baselines in view of automated driving tasks.

The first metric evaluates the depth accuracy and is defined as the percentage of the disparity estimates that are considered inliers [55]. A disparity estimation with an absolute deviation of less than or equal to 3 px and a relative deviation less than 5 % compared to ground truth is considered as an inlier. The second metric assesses the semantic performance and is defined as the average Intersection-over-Union (IoU) over all classes [56]. Third and fourth, we report the runtime and use the number of Stixels per image as a proxy to assess the complexity of the obtained representation. The runtime is obtained using an FPGA for SGM, a GPU (NVIDIA Titan X) for the FCN, and a CPU (Intel Xeon, 10 cores, 3 GHz) for the Stixel segmentation. Note that a system suitable for autonomous driving is expected to reach excellent performance in all four metrics simultaneously.

### 7.3. Baseline

As a reference model, we perform what we call *smart downsampling*, since regular downsampling of the disparity image would result in very strong blocking artifacts, in particular on the ground plane. Instead, we use the pixel-level semantic input to differentiate three depth models, analog to the three structural classes used in Stixels. For ground pixels, we assign the mean deviation to the flat ground hypothesis; for pixels covering vertical obstacles, we assign the mean disparity; and for sky pixels we assign disparity zero. The downsampling factor is chosen such that the number of bytes required to encode this representation is identical to the number of bytes required to encode the number of Stixels that we compare with. For example, 700 Stixels as are typical for KITTI are equivalent to a downsampling factor of 21 in the smart downsampling method. Note that while such a downsampling can serve as a compression method, it lacks robustness and does not provide a medium-level representation as required by subsequent processing stages [8, 9, 10, 11, 12, 13, 14, 15].

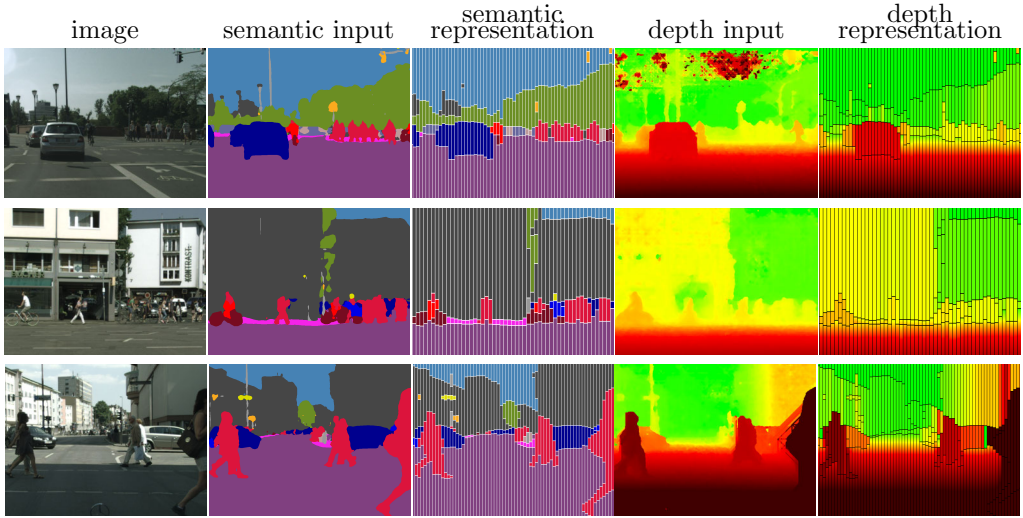

Figure 6: Example output of our Stixel segmentation in terms of semantic and depth representation on Cityscapes [20] with color encodings as in Fig. 1. Even objects as small as traffic lights and signs are represented accurately.

*7.4. Stixel Model*

One of the central objectives of the Stixel representation is to compress the content of the scene to a small number of values with as few computational overhead as possible, *c.f.* Fig. 6. In the first experiment, we therefore focus on evaluating the core Stixel parameters that influence the number of resulting Stixels and the computation time: the horizontal extent or downscaling factor $w_s$, which controls the discretization; the vertical downscaling factor $h_s$, which has a quadratic influence on the execution time; and the model complexity term $\Psi_{\mathrm{mc}}$, which controls the number of Stixels per image column. Figure 7 (bottom) shows individual sweeps of these values, intersecting at our base parameter configuration. It is interesting to see that sweeping $h_s$ has the strongest influence on both depth and semantic accuracy, while the other two parameters have a stronger influence mainly on the number of Stixels.

To evaluate the general Stixel model assumptions, irrespective of the quality of input data, we compute Stixels where both estimated disparity and semantic input are replaced with ground-truth data. By comparing the resulting Stixels with exactly this ground truth again, we are able to assess the loss induced by imposing the Stixel model onto a scene. The results of this experiment are shown in Fig. 7 (top). Despite the strong compression of the scene information, we are able to retain roughly 94 % disparity and 85 % semantic accuracy with our base parameterization, compared to 91.8 % and 76.7 %, respectively, for our smart downsampling at equal compression rate.

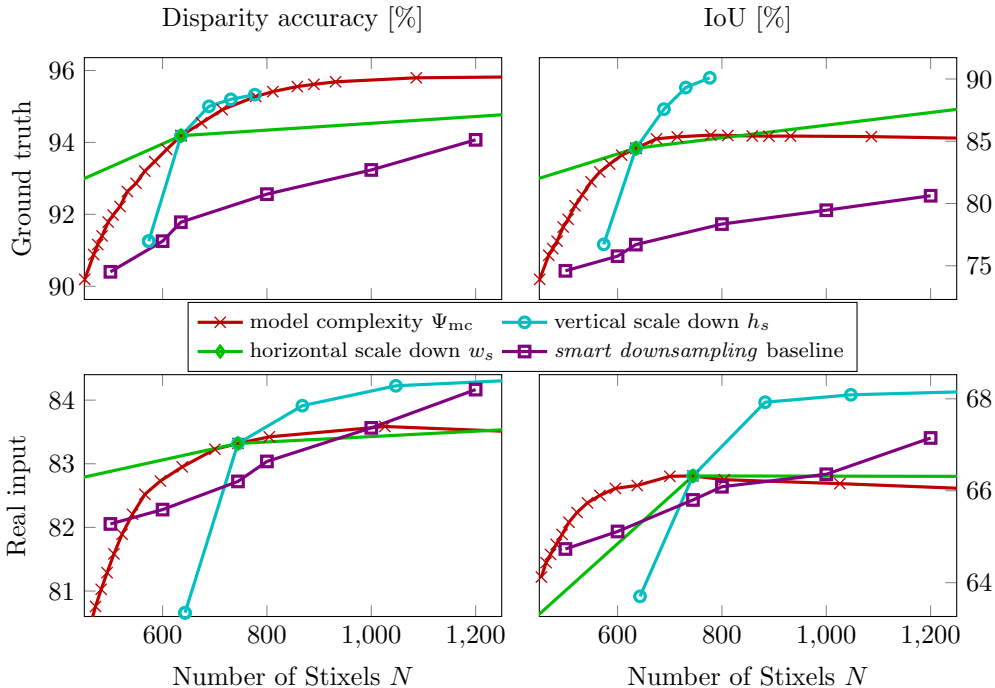

Figure 7: Influence of important model parameters on disparity and semantic accuracy as well as the number of Stixels. Results are reported on the Ladicky dataset [54]. The results from the top row are obtained using ground-truth depth and semantics as input and naturally serve as upper bound for the results of our full system shown on the bottom. As a reference, we propose a *smart downsampling* baseline. See text for more details.

Another central aspect of the Stixel model we put forward in this paper is the use of prior information on the scene structure. In the next experiment, we evaluate the effect of $\Psi_{\text{grav}}$ and $\Psi_{\text{do}}$ in Eq. (10), which control gravitational as well as depth ordering constraints. We find that these priors have little influence when facing only fair weather conditions as in [53, 55, 20]. As prior information is most valuable in situations where input data is weak, we demonstrate the benefit of these terms on a dataset with bad weather conditions [4], where many false positive Stixels are detected on the road due to missing or false depth measurements. Figure 8 shows ROC curves that are obtained by varying the model complexity $\Psi_{\text{mc}}$ and evaluating by following to the dataset protocol. We repeat the experiments once with and without our structural priors enabled, which clearly indicates how the prior information helps to reduce the number of false positives while keeping the detection rate high.

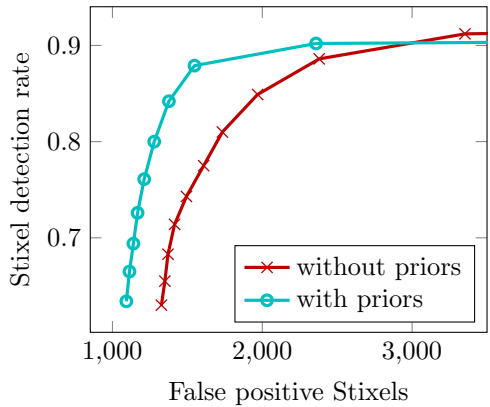

Figure 8: Stixel detection and false positive rates in bad weather scenarios [4]: the structural priors, *c.f.* Section 4.1, reduce the number of false alarms caused by weak input data.

*7.5. Impact of Input Cues*

The Stixel inference takes into account data likelihoods from three different sources: disparity data $\boldsymbol{D}_:$, color information $\boldsymbol{I}_:$, and pixel-level semantic label scores $\boldsymbol{L}_:$. To assess the importance of different input modalities, we report results when using only a single modality as input, as well as all combinations of them. Note that when depth or semantic labels are removed as input channel, we still assign this information in a post-processing step to be able to evaluate all metrics. We provide qualitative results in Fig. 9 and Table 1 summarizes this experiment, where it is interesting to see that adding color or semantics helps to improve depth accuracy, and adding color or depth helps to improve semantic accuracy. However, we also see that the gain of adding color information is only minor, in particular if semantic labels are available and provide a powerful representation of the appearance channel. Further, increased accuracy coincides with a slightly larger number of Stixels, as indicated in the last row of Table 1. The smart downsampling baseline performs surprisingly well on Ladicky [54] and KITTI'15 [55] given that its output resolution is $58 \times 18$ pixels. On Cityscapes [20], where ground-truth annotations are finer and more precise, the gap is slightly larger, *i.e.* 54.1 % compared to 60.1 % IoU. However, as evident from the qualitative results in Fig. 9, the differences in terms of representational richness are much larger than captured by the metrics.

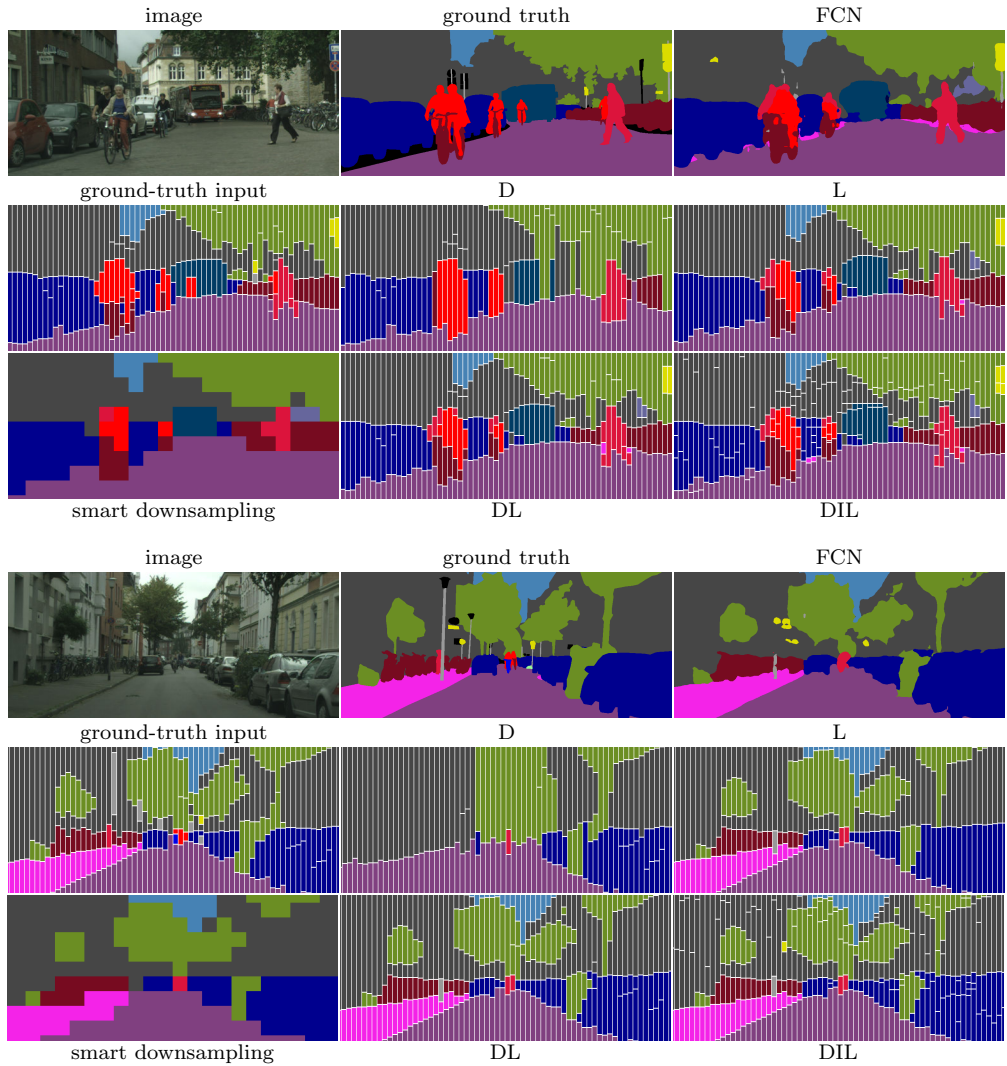

Figure 9: Qualitative segmentation and labeling results of the proposed method with different combinations of input modalities (**D**epth, Color **I**mage, Semantic **L**abels) compared to groundtruth, the raw FCN input, and our *smart downsampling* baseline. Note that objects of the same class, *e.g.* vehicles in the front, are not separated if stereo information is missing (L). Without semantic input (D), important objects, *e.g.* sidewalk (bottom) and the bus in farther distance (top), are not recognized. The color information yields an over-segmentation of the image and helps to detect small objects, *e.g.* traffic light (bottom) as well as object boundaries more precisely. The smart downsampling discards important smaller objects like traffic signs; however, these have a minor impact in the pixel-level metrics in the presented qualitative results in Table 1.

Table 1: Results of the proposed method with different combinations of input modalities (**D**epth, Color **I**mage, Semantic **L**abels) compared to the raw inputs (SGM and FCN) and our *smart downsampling* baseline (SDS). We evaluate on three different datasets (Ladicky [54], KITTI'15 [55], and Cityscapes [20]) using four metrics, *c.f.* Sections 7.1 and 7.2. Note that the runtime is reported as the sum of all individual components. However, in order to maximize the throughput of the system, SGM and FCN can be computed in parallel to the Stixel algorithm with the cost of one frame delay. Doing so yields a frame rate of the DIL variant of 47.6 Hz on KITTI and 15.4 Hz on Cityscapes.

| Metric | Data | SGM FCN | SDS | D | I | L | DI | DL | IL | DIL |
|---|---|---|---|---|---|---|---|---|---|---|
| Disp. acc. [%] | [54] | 82.4 | 82.7 | 80.0 | 47.9 | 72.7 | 80.9 | 83.0 | 72.6 | **83.3** |
|  | [55] | 90.6 | 88.9 | 90.4 | 58.7 | 77.5 | 90.7 | 91.3 | 81.9 | **91.4** |
| IoU [%] | [54] | **69.8** | 65.8 | 46.1 | 27.2 | 62.5 | 47.3 | 66.1 | 66.8 | 66.5 |
|  | [20] | **60.8** | 54.1 | 43.8 | 18.1 | 59.7 | 44.2 | 60.0 | 59.8 | 60.1 |
| Runtime [ms] | [54] | 39.2 | — | 24.2 | 2.7 | 25.5 | 24.7 | 45.7 | 25.7 | 46.6 |
|  | [20] | 110 | — | 70.5 | 10.5 | 86.1 | 70 | 143 | 86.4 | 148 |
| No. of Stixels | [53] | 0.5 M[a] | 745[b] | 509 | 379 | 453 | 652 | 625 | 577 | 745 |
|  | [20] | 2 M[a] | 1395[b] | 1131 | 1254 | 1048 | 1444 | 1382 | 1283 | 1395 |

[a] We list the number of pixels for SGM and FCN raw data to approximately compare the complexity to Stixels.

[b] We use a downsampling factor that results in the same number of bytes to encode this representation as for the given number of Stixels.

Given that depth and semantics are the two major input cues, we now take a closer look into the balance of those two modalities by varying the semantic score weight $\delta_L$ from Eq. (19) while keeping $\delta_D$ from Eq. (15) fixed to 1. Fig. 10 shows the disparity accuracy on the Ladicky [54] and KITTI'15 [55] datasets together with the semantic IoU scores on the Ladicky [54] and Cityscapes [20] dataset, plotted over the semantic score weight $\delta_L$. We conduct the experiment twice, once for a Stixel width 2, which is slow to compute but yields high accuracy, and once for our standard width 8. It can be seen that increasing the influence of semantics gradually reduces the number of disparity outliers, with a weight of 5 being the best trade-off between both measures. It is interesting to see, however, that a too strong influence can also have the inverse effect, where a large semantic weight actually decreases the semantic performance and increases the number of disparity outliers, $c.f.$ Fig. 10 (bottom left). In these cases, the Stixel model can act as a regularizer and yield better precision than the pixel-level input.

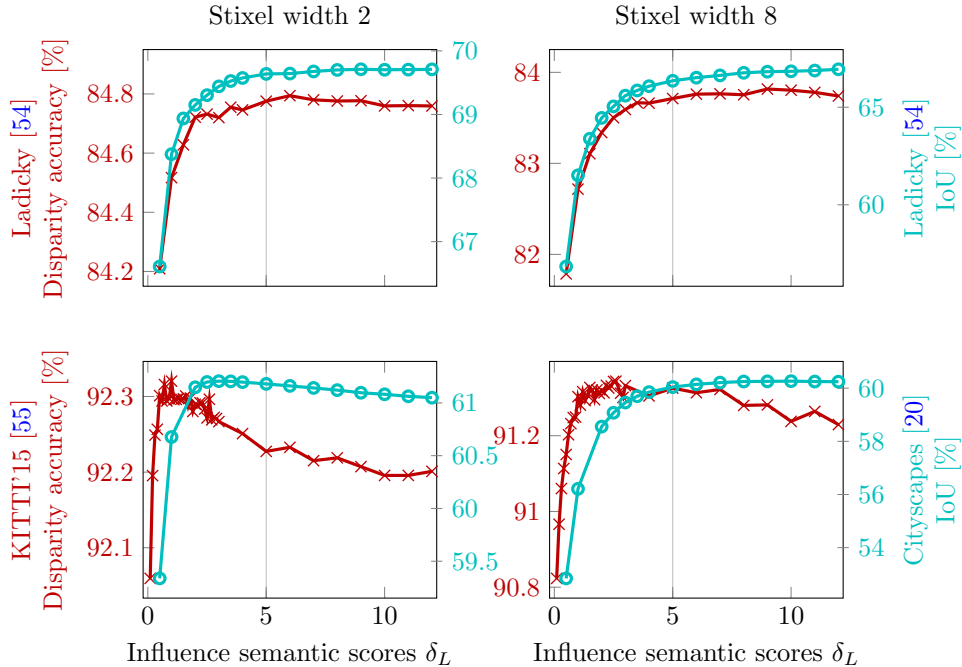

Figure 10: Analysis of the influence of the semantic scores $\delta_L$. We evaluate Stixels with width 2 (left column) and width 8 (right column) regarding four metrics: (1) Disparity accuracy on Ladicky [54] (top row, red), (2) IoU on Ladicky [54] (top row, blue), (3) Disparity accuracy on KITTI'15 [55] (bottom row, red), and (4) IoU on Cityscapes [20] (bottom row, blue).

## 7.6. Approximations

In Section 5.3, we introduced several approximations during inference to keep asymptotic and practical run time low. In our final experiment, we compare these approximations with an exact inference of our model on Ladicky [54], where the number and size of the images is small enough to conduct this experiment. As is evident from Table 2, both variants yield the same results quality, supporting the validity of our approximations. However,

Table 2: Performance of our approximated inference, *c.f.* Section 5.3, compared to an exact inference of Eq. (20). While the accuracy of both variants is identical, the runtime of our inference scheme is two orders of magnitude lower.

|  | our approximated inference | exact inference |
|---|---|---|
| Disparity accuracy [%] | 83.3 | 83.1 |
| IoU [%] | 66.5 | 66.3 |
| No. of Stixels per image | 745 | 741 |
| Inference run time [ms] | 7.4 | 863 |

the run time of the approximate Stixel inference is two orders of magnitude lower.

## 8. Discussion and Conclusion

In this paper, we have presented the Stixel World, a medium-level representation of image and depth data with a particular focus on automotive vision in complex traffic environments. We claim that traffic scenes are structured mostly in the vertical domain and can be modeled in terms of horizontal (ground) and vertical (obstacles) planar surfaces. From our experiments, in particular from Figs. 5 and 7 and Table 1, we observe that our basic model assumptions are valid and indeed applicable to adequately represent complex traffic scenes. Further, our Stixel model is designed to serve as a primitive structuring element for a wide range of automotive vision applications by taking into account and benefiting from as many sources of information as possible. Our results indicate that our Stixel formulation indeed combines multiple input modalities such that they complement each other, *c.f.* Table 1 and Fig. 10. Another important property of the Stixel World is compactness. We have demonstrated that there is a loss in representation accuracy induced by the compression properties of the Stixel model. However, we are able to retain approximately 94% disparity and 85% semantic accuracy, which is significantly better than the smart compression variant we have compared our model against. Note that the loss in semantic accuracy does not stem from a confusion of different object classes, but almost exclusively results from Stixel discretization artifacts at object boundaries that significantly impact the underlying pixel-wise Intersection-over-Union (IoU) metric, *c.f.* Fig. 6.

It is instructive to place the Stixel World model, its properties, and the performance obtained in context by comparing what would be necessary in a realistic application in the area of automotive vision. The constant trend towards higher camera resolutions puts a significant computational burden on in-vehicle processing units. For reference, in the upcoming years, the number of image and 3D points in real-world applications is expected to increase to several million. This calls for an efficient model to abstract from the vast amount of raw data. We feel that the Stixel World is an ideal medium-level representation given the properties demonstrated in this paper. Our full Stixel model including all input cues can be computed in real-time frame rates on automotive-grade hardware. However, finding the right parameter set that is suitable for all applications remains a difficult task. This open issue motivated our investigation into automatic parameter learning via structured SVM models, as shown in Section 6, yet with limited success. Still, we feel that these techniques should be further investigated in future work. Furthermore, an extension to a system with an online estimation of parameters to adapt for changing environment conditions is desirable.